\documentclass[letterpaper]{article} % DO NOT CHANGE THIS
\usepackage[]{aaai24}  % DO NOT CHANGE THIS
\usepackage{times}  % DO NOT CHANGE THIS
\usepackage{helvet}  % DO NOT CHANGE THIS
\usepackage{courier}  % DO NOT CHANGE THIS
\usepackage[hyphens]{url}  % DO NOT CHANGE THIS
\usepackage{graphicx} % DO NOT CHANGE THIS
\usepackage{natbib}  % DO NOT CHANGE THIS AND DO NOT ADD ANY OPTIONS TO IT
\usepackage{caption} % DO NOT CHANGE THIS AND DO NOT ADD ANY OPTIONS TO IT
\usepackage{algorithm}
\usepackage{algorithmic}

\usepackage{newfloat}
\usepackage{listings}
\DeclareCaptionStyle{ruled}{labelfont=normalfont,labelsep=colon,strut=off} % DO NOT CHANGE THIS
\floatstyle{ruled}
\newfloat{listing}{tb}{lst}{}
\floatname{listing}{Listing}
\usepackage{diagbox}
\usepackage{amsthm,amsmath,amssymb}
\usepackage{mathrsfs}
\usepackage{multirow}
\usepackage{subfigure}
\usepackage{makecell}
\newcommand{\DataAugmentation}[1]{Data Transformation}
\newcommand{\dataaugmentation}[1]{data transformation}
\newcommand{\Dataaugmentation}[1]{Data transformation}
\title{\DataAugmentation{} to Construct a Dataset for Generating Entity-Relationship Model from Natural Language}
\author{
    Zhenwen Li\textsuperscript{\rm 1},
    Jian-Guang Lou\textsuperscript{\rm 2},
    Tao Xie\textsuperscript{\rm 1}\thanks{~~The corresponding author is Tao Xie.}
}
\affiliations{
    \textsuperscript{\rm 1}School of Computer Science, Peking University, China;\\
    \textsuperscript{\rm 2}Microsoft Research Asia, Beijing, China;\\
    lizhenwen@pku.edu.cn,
    jlou@microsoft.com,
    taoxie@pku.edu.cn
}

\usepackage{bibentry}
\begin{document}

\maketitle

\begin{abstract}
In order to reduce the manual cost of designing ER models, recent approaches have been proposed to address the task of NL2ERM, i.e., automatically generating entity-relationship (ER) models from natural language (NL) utterances such as software requirements. These approaches are typically rule-based ones, which rely on rigid heuristic rules; these approaches cannot generalize well to various linguistic ways of describing the same requirement. Despite having better generalization capability than rule-based approaches, deep-learning-based models are lacking for NL2ERM due to lacking a large-scale dataset. To address this issue, in this paper, we report our insight that there exists a high similarity between the task of NL2ERM and the increasingly popular task of text-to-SQL, and propose a \dataaugmentation{} algorithm that transforms the existing data of text-to-SQL into the data of NL2ERM.
%Our data transformation scripts are publicly available at https://github.com/lzw-pku/NL2ERM-data-transformation.
We apply our \dataaugmentation{} algorithm on Spider, one of the most popular text-to-SQL datasets, and we also collect some data entries with different NL types, to obtain a large-scale NL2ERM dataset.%\footnote{We will release our \dataaugmentation{} script and the dataset at the publication of this paper; we currently do not release them due to the double blind requirement.}.
Because NL2ERM can be seen as a special information extraction (IE) task, we train two state-of-the-art IE models on our dataset.
The experimental results show that both the two models achieve high performance and outperform existing baselines.
\end{abstract}
\section{Introduction}
Data modeling is the task to design a data model underlying a system under development.
A commonly used type of data model is an entity-relationship (ER) model, which can be of great help in designing a database. An ER model consists of entities, attributes of entities, and relationships. 
There are two types of relationships: the relationship between a pair of entities (entity-entity relationship), and the relationship between an entity and an attribute (entity-attribute relationship).
Given that manually designing an ER model is a daunting task~\cite{heuristic-based}, recent approaches~\cite{Novel-Natural-Language, machine-learning, Generating-ER-Diagrams, heuristic-based, Scenario-based} use two steps to automatically generate ER models from natural language (NL) utterances such as software requirements, by first extracting entities/attributes and then extracting their relationships. 

The preceding existing NL2ERM approaches (typically rule-based ones that rely on rigid heuristic rules)  face two major limitations, causing them not to generalize well to various linguistic ways of describing the same requirement. 
First, a large number of rules are needed to cover the possible \textbf{synonyms} of a word appearing in the NL utterances.  
For example, there is a widely used rule: \textit{if two nouns are separated by ``have'' or ``has'',  then the former is an entity, while the latter is an attribute} \cite{Scenario-based, Generating-ER-Diagrams}.
However, there are a lot of sentences that the rule cannot process because the sentences use synonyms of the verb ``have'' (such as ``own'', ``contain'', and ``possess'') not covered by the rule.
Second, a large number of rules are needed to cover the possible \textbf{patterns} of NL utterances, because these rules process the NL utterances based on templates or regular expressions, instead of semantics.
For example, there is a  widely used rule: \textit{consecutive nouns separated by commas or ``and'' are attributes} \cite{Novel-Natural-Language, Scenario-based}.
However, the sentence ``Students are described by name and age'' can be written as ``Students not only have the name, but have the age as well'', and the latter pattern of the sentence is not applicable to the rule.

Compared with rule-based approaches, deep-learning-based NL2ERM models can provide better generalization capability  for various tasks~\cite{nmt_sota, nl2sql_sota, relation_prediction_sota}; however, there are two main reasons for lacking a large-scale dataset (needed by these models), which is not readily available from previous related work~\cite{Generating-ER-Diagrams, Novel-Natural-Language,heuristic-based}.
First, there are only a small number of data entries\footnote{A data entry in NL2ERM includes  some NL utterances and the corresponding ER model.} in the datasets from the previous work, and there is no publicly available dataset released by the authors of the previous work.
Most of the previous work evaluates their proposed approaches based on only case studies~\cite{Generating-ER-Diagrams, Novel-Natural-Language}.
One exception is ER-Converter \cite{heuristic-based}, which is evaluated on a dataset with 30 entries, but the dataset is not publicly released.
Second, training a deep-learning-based model needs data entries with fine-grained annotations (not provided by the previous work).
A token-level annotation can greatly improve the model's performance.
The data used in the previous work does not annotate which tokens in utterances correspond to specific entities/attributes.

To construct a large-scale fine-grained annotated dataset with reasonable cost, our insight is that the existing data of text-to-SQL \cite{rat_sql, irnet} can be transformed into the data of NL2ERM.
A data entry of text-to-SQL includes three parts: an NL utterance, a SQL query, and a corresponding database.
Note that as a well-studied task, text-to-SQL has many datasets \cite{yu-etal-2018-spider, wikisql, unimer_benchmark}. For example, the Spider dataset~\cite{yu-etal-2018-spider} contains 200 databases and 10181 $<$NL utterance, SQL query$>$ pairs, and the wikisql dataset~\cite{wikisql} contains 24241 databases and 87726 $<$NL utterance, SQL query$>$ pairs. 

\begin{figure}[t]
\centering 
\includegraphics[width=0.5\textwidth]{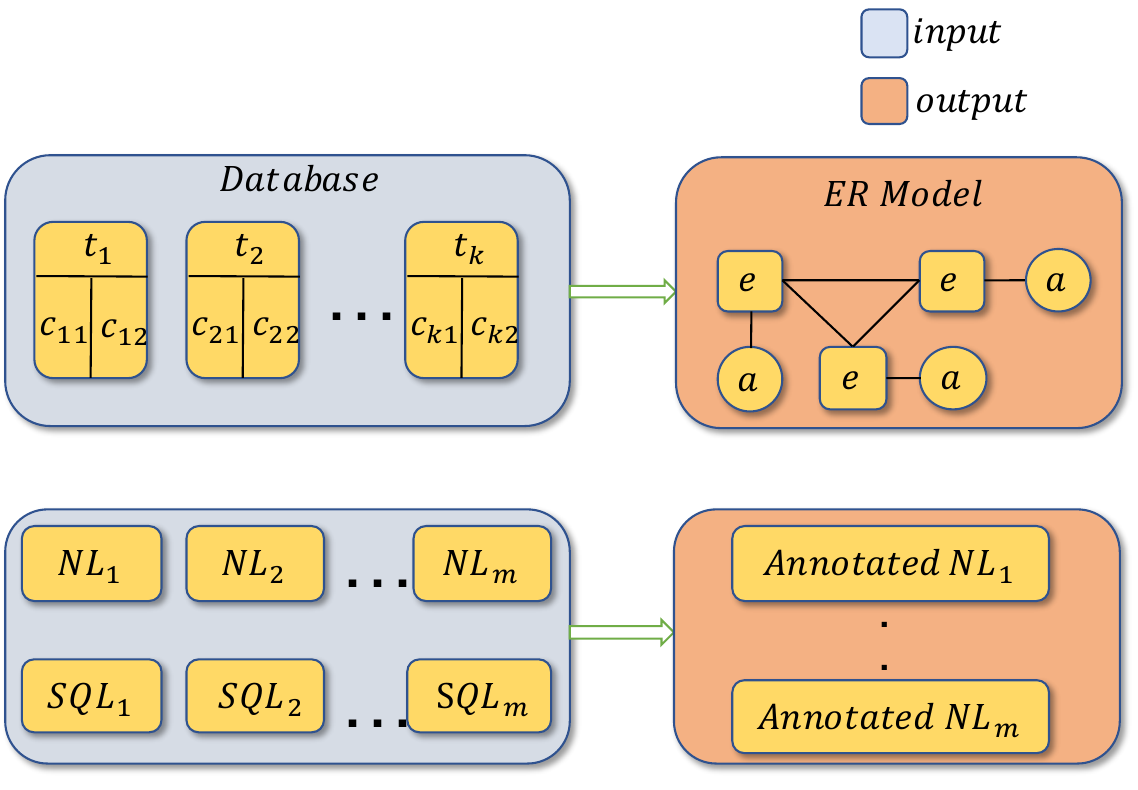}
\caption{An overview of the correspondence between the data of text-to-SQL and the data of NL2ERM.}
\label{fig:intro}
\end{figure}

There is a correspondence between the data entry of text-to-SQL and NL2ERM.
We illustrate this correspondence in Figure \ref{fig:intro}.
First, a  database can be transformed into an ER model because there is a high correlation between them \cite{Database2ERM}.
The tables, columns, and foreign keys in the database typically correspond to the entities, attributes, and entity-entity relationships in the ER model.
Second, the NL utterances in text-to-SQL data can be used as the input of NL2ERM because these NL utterances contain information about the database/ER model.
Third, the SQL queries correspond to fine-grained annotations in NL2ERM. We can automatically get the tables (corresponding to entities) and columns (corresponding to attributes) that appear in the NL utterances, so that we can obtain the annotation on which entities/attributes appear in the NL utterances.

Based on this insight, in this paper, we propose a novel \dataaugmentation{} algorithm to transform the existing data of text-to-SQL into the data of NL2ERM in three steps.
The input to our algorithm is a database and the corresponding $<$NL utterance, SQL query$>$ pairs.
First, we transform the database into a raw ER model.
Second, we apply schema linking between NL utterances and corresponding SQL queries, and generate token-level annotations based on the raw ER model and the schema linking results.
Third, we prune the entities and attributes that do not appear in NL utterances of the raw ER model to get the final ER model based on the schema linking results.

We apply the \dataaugmentation{} algorithm on Spider \cite{yu-etal-2018-spider}, one of the most popular text-to-SQL datasets, to obtain a large-scale NL2ERM dataset, which contains 160 ER models, each corresponding to 50 sentences on average. We split this dataset into a training dataset (containing 140 data entries) and a test dataset (containing 20 data entries) according to the split of Spider. 

Because the type of NL utterances in text-to-SQL is  question, we also collect some data entries with different NL types (software requirement and scenario description) as a supplement to the test dataset. 
The final test dataset consists of three parts according to the types of NL utterances: the  question part, requirement part, and scenario part.

Because NL2ERM can be seen as a special information extraction (IE) task, we train two state-of-the-art IE models, UIE \cite{uie} and REBEL \cite{rebel}, on our dataset.
An IE model can extract entities with different entity types and the relationships between entities from the given NL utterances.
To apply IE models on NL2ERM, we treat both entities and attributes in NL2ERM as entities in IE with entity types \textit{Entity} and \textit{Attribute}, respectively.
We also treat the entity-entity relationship and entity-attribute relationship in NL2ERM as two relationships in IE.

We conduct experiments on training and evaluating the IE models on our datasets.
The experimental results show that the two deep-learning-based IE models achieve high performance and outperform existing baselines.

In summary, this paper makes the following main contributions:
\begin{itemize}
    \item We propose a novel \dataaugmentation{} algorithm to transform the data of text-to-SQL into the data of NL2ERM.
    \item We apply our \dataaugmentation{} algorithm on Spider (one of the most popular text-to-SQL datasets), and collect some data entries with different NL types to produce a dataset, which is the first large-scale fine-grained dataset for NL2ERM, to the best of our knowledge.
    \item The experimental results show that our dataset can help train deep-learning-based models on NL2ERM.
\end{itemize}
\section{Related Work}
In this section, we first introduce previous work in NL2ERM, and then discuss the relationship and differences between NL2ERM and information extraction (IE).

\noindent \textbf{NL2ERM}. 
Most research efforts in this field of NL2ERM propose rule-based approaches  \cite{EER, GOME_ERM, Omar_ERM, chen_ERM}. 
For example, ER-converter \cite{heuristic-based} is a rule-based approach to generate ER models from NL utterances.
ER-converter is not an automatic approach, and human intervention is needed.
\citet{Generating-ER-Diagrams} propose a rule-based approach named ER generator, which improves the rules in ER-converter by including  multiple new rules.
\citet{Scenario-based} propose a rule-based approach that generates ER models from NL  utterances and translates NL utterances into SQL.
\citet{Novel-Natural-Language} propose an  approach named ER-Gen, which utilizes multiple NLP techniques such as sentence segmentation and POS tagging.
Although these approaches include  different heuristic rules to extract ER models from NL utterances, these rules are too rigid and cannot be generalized well to various linguistic ways of describing the same requirement.
\citet{machine-learning} propose a machine-learning model that uses four types of classifier: Random Forest, Naive Bayes, Decision Table, and SMO. 
But this machine learning model can process only software descriptions and cannot extract relationships.

\noindent\textbf{IE}. 
The goal of IE is to extract (from NL utterances) relational tuples, each of which is in the format of $<entity_1, relation$ $type, entity_2>$.
Some research efforts in IE treat entity extraction as a sequence labeling task~\cite{IE_label_1, IE_label_2, IE_label_3}. 
They annotate  tokens with different labels, e.g., they annotate  entity-corresponding tokens with  their entity types.
As for relationship extraction, \citet{relation_extraction_1} propose a relationship extraction model that inserts four reserved word pieces to mark the beginning and the end of each entity mentioned in a sentence, and their model achieves high performance. 
\citet{uie}  propose a universal generation model to generate a structured extraction language to represent  relationships.

NL2ERM is similar to the IE  task~\cite{IE_survey}.
Entities and attributes in NL2ERM can be seen as entities in IE with two different entity types, while the relationships in NL2ERM can be seen as relationships  with different relation types in IE.
The similarity between the two tasks inspires us to train IE models on our NL2ERM dataset to produce  NL2ERM models.

Despite the similarity, there are two major differences between NL2ERM and IE, so existing IE models that are trained on IE datasets cannot be directly used for the NL2ERM task.
First, an ``entity'' or ``attribute'' in NL2ERM is usually a common noun or concept, whereas an ``entity'' in IE can be a proper noun such as ``sun'' and ``America'', and there is no entity type ``attribute'' in IE.
%Second, the natural language in NL2ERM usually describes a system under development, whereas the natural language in IE has no such constraint.
Second, a ``relationship'' in NL2ERM can be between two entities or an entity and an attribute, whereas a ``relationship'' in IE must be between two entities.

\section{\DataAugmentation{}}
In this section, we introduce the detail of our \dataaugmentation{} algorithm.
\subsection{Overview}
We formulate the \dataaugmentation{} process as follows.

\textbf{Input}. The input consists of two parts: a database and a set of $<$NL utterance, SQL query$>$ pairs.
The database $D$ contains $k$ tables $\{t_1, t_2, .., t_k\}$, and each table $t_i$ has multiple columns $c_{ij}$.
The set of $<$NL utterance, SQL query$>$ pairs is represented as $\{<n_1, s_1>, <n_2, s_2>, .. <n_m, s_m>\}$.

\textbf{Output}. The output also consists of two parts: a set of $<$NL utterance, Entity, Attribute, Relationship$>$ tuples and an ER model.
The former part is the fine-grained annotations of the input NL utterance.
The annotations are represented as $\{<l_1, e_1, a_1, r_1>, <l_2, e_2, a_2, r_2>, .., <l_m, e_m, a_m,$ $ r_m>\}$, where $l_i$ is a token-level annotation to annotate entity/attribute  tokens, and $e_i$, $a_i$, and $r_i$ are the sets of entities, attributes, and relationships in the NL utterance, respectively. 
The ER model is composed of three sets: the entity set $E$, the attribute set $A$, and the relationship set $R$, where $E$, $A$, and $R$ are the union sets of $e_i$, $a_i$, and $r_i$, respectively.
The token-level annotation will be used to train deep-leaning-based models, and the ER model is the ground truth in evaluation.
\begin{figure*}[!t]
\centering 
\includegraphics[width=1\textwidth]{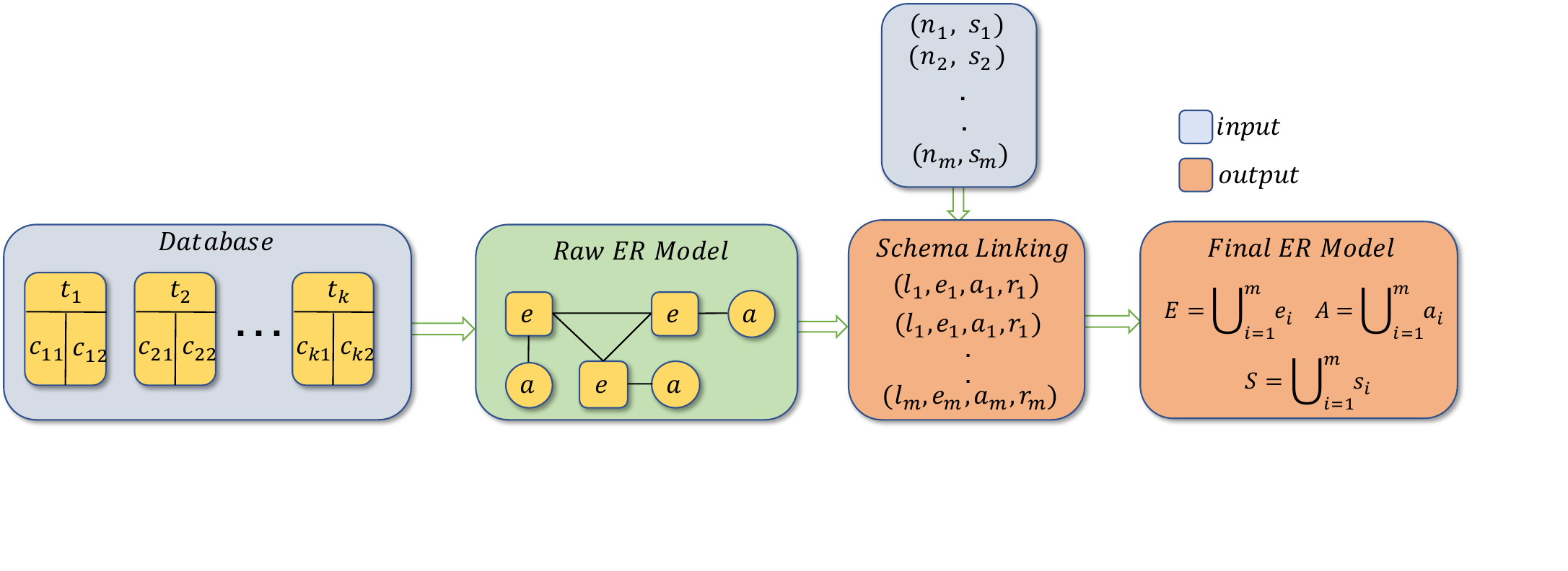}
\caption{An overview of the \dataaugmentation{} process}
\label{fig:synthesizer}
\end{figure*}

Figure \ref{fig:synthesizer} illustrates the overview of our \dataaugmentation{} process. 
For each database in the dataset of text-to-SQL, we collect $<$NL utterance, SQL query$>$ pairs that correspond to the database from the dataset, and then the \dataaugmentation{} algorithm generates the data of NL2ERM in three steps: transforming the database into a raw ER model, conducting schema linking between NL utterances and corresponding SQL queries, and generating the final ER model. 
The motivation for schema linking is that the results show which tokens in NL utterances correspond to the entities and attributes. 
The difference between the raw ER model and the final ER model is that the final ER model contains only the entities/attributes and relationships that appear in the NL utterances, while the raw ER model may contain those that do not appear.
%The reason to prune the raw ER model is some tables and columns in the database are not covered by natural language queries, and these should be eliminated from the raw ER model.
\begin{figure}[!t]
\centering 
\includegraphics[width=0.5\textwidth]{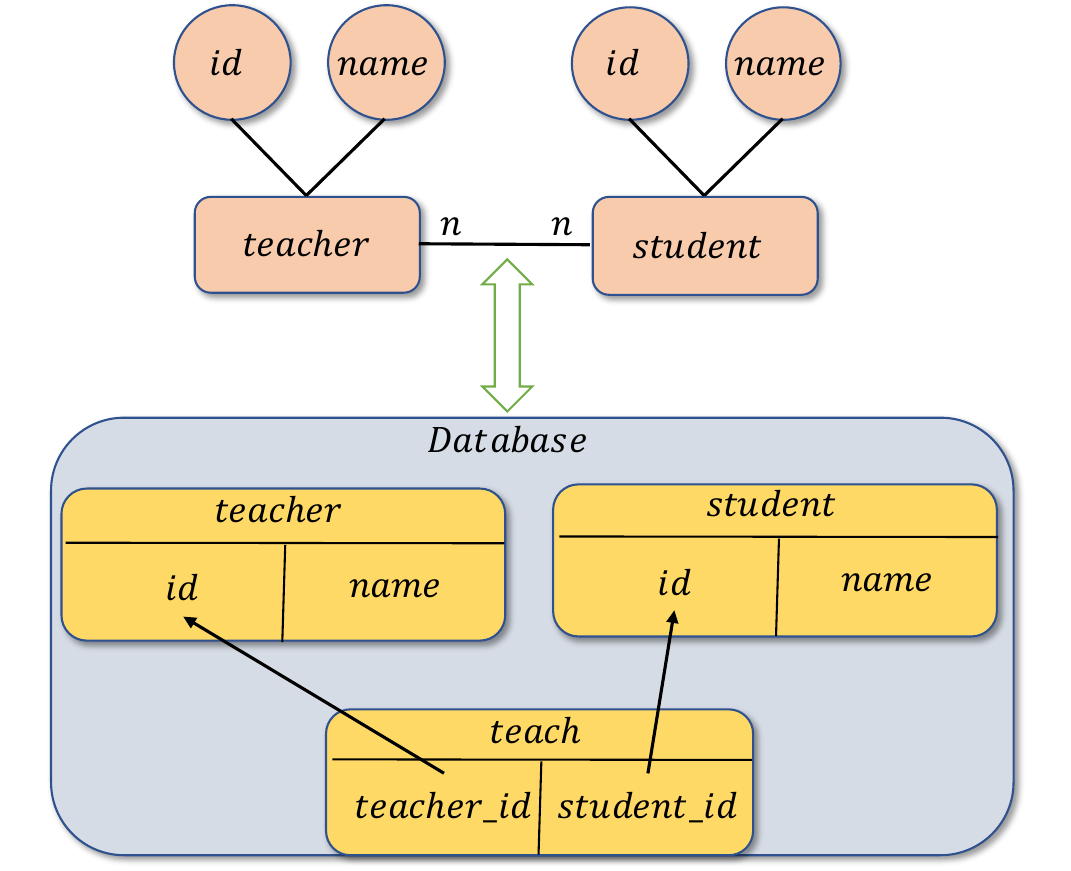}
\caption{An example of a special case: the relationship between \textit{Teacher} and \textit{Student} has a cardinality of many-to-many, so the relationship corresponds to a table (\textit{teach}) in the database.}
\label{fig:special_case}
\end{figure}
\subsection{Transformation of Database into Raw ER Model}
There is a correspondence between ER models and databases.
Typically, the entities, attributes, and entity-entity relationships in an ER model correspond to the tables, columns, and foreign keys in a database, respectively, and an entity-attribute relationship corresponds to a relationship that a table contains a column.

However, an entity-entity relationship in the ER model can correspond to a table rather than a foreign key in the database in a special case.
We name such tables that correspond to an entity-entity relationship as \textit{relation table}.
As shown in Figure \ref{fig:special_case}, the special case is that the entity-entity relationship has a cardinality of many-to-many (an instance of one entity can correspond to many instances of the other entity, and vice versa), or only part of both entities participates in the relationship.
Due to this special case, simply transforming databases according to the correspondence can lead to an inaccurate ER model.

We find that the \textit{relation tables} typically contain exactly two columns, each of which has a foreign key connection with other tables, and sometimes the \textit{relation tables} have an extra column as the primary key of the table.
Based on our observation, we apply an algorithm to find the \textit{relation tables} in the database.
First, we count the number of foreign keys and the number of columns in each table.
Second, we determine  \textit{relation tables} as  the tables that contain no more than three columns and exactly two foreign keys.

After finding all the \textit{relation tables} in the database, we transform the remaining part of the database into an ER model according to the preceding correspondences, and add the \textit{relation tables} into the ER model as entity-entity relationships.
\subsection{Schema Linking}
Schema linking is a widely used technique in text-to-SQL \cite{rat_sql, irnet}.
The goal of schema linking is to recognize which tokens in the NL utterances correspond to tables and columns.
Previous work~\cite{rat_sql, irnet} typically uses string-match to match the tokens of the NL utterances with the names of tables/columns.

For our work, the goal of schema linking is to recognize the entities, attributes, and relationships in an NL utterance.
There are some differences between the previous work and ours.
In text-to-SQL, the target SQL query is invisible to the model, so the schema linking algorithm needs to match the tokens with all the tables/columns in the database.
Our \dataaugmentation{} algorithm takes advantage of the corresponding SQL query, which contains all potential tables/columns that may appear in the NL utterance.
In this way, the \dataaugmentation{} algorithm needs to match the tokens with only the tables/columns that appear in the corresponding SQL query, and the range of matches has been greatly narrowed.

Our schema linking consists of three steps.
First, we extract the tables/columns from the SQL query. 
To further narrow the range, we remove \textit{relation tables} and their columns  because these tables/columns do not correspond to any entities/attributes in the raw ER model.
Second, we enumerate all n-grams of lengths 1 to 5 in the NL utterance in the descending order of length, and we match each of them  with the candidate tables/columns names.
If an n-gram exactly matches the name of a table/column, or it is a sub-sequence of the name of a table/column, we label the tokens in this n-gram as \textit{Table}/\textit{Column}.
Once an n-gram is recognized, we remove other n-grams that overlap with it from the candidates, and doing so ensures that each token has at most one type of label.
After string-matching, we label all the unlabeled tokens as \textit{Other}.
Third, we map all of the matched tables/columns  into the raw ER model to get the corresponding entities/attributes. 
For all relationships between the matched tables/columns, if the relationship appears in the SQL query, such as table joining, or belonging relationship between tables and columns, we also map the relationship into the raw ER model to get the corresponding relationship.

After schema linking, we get a token-level annotation for each NL utterance, we can also get the set of entities, attributes, and relationships that appear in each NL utterance.
\subsection{Generating Final ER Model}
For a database in text-to-SQL, some tables/columns never appear in any corresponding NL utterances.
So the transformed raw ER model cannot be directly used as the ground truth ER model for NL2ERM.
We prune the raw ER model by deleting the unused tables/columns to produce the final ER model.
The final ER model is composed of three sets: the entity set, attribute set, and relationship set; each of the sets is a union set of the corresponding set of each NL utterance.
\section{Dataset}
\begin{table*}[!th]
  \centering
    \begin{tabular}{l|ccccccc}
        \hline
        ~&ER model&utterance&avg length&entity&attribute&relationship\\
        \hline
        Train dataset&140&7000&14.2&8068&10883&15810\\
         Question part&20&1034&13.8&1103&1431&2022\\
         Requirement part&2&18&8.7&28&24&31\\
         Scenario part&6&37&17.1&52&80&98\\
        \hline
    \end{tabular}
    \caption{Statistical results of our dataset, where ``avg length'' indicates  the average number of tokens in an utterance.}
    \label{tab:dataset}
\end{table*}
\subsection{\DataAugmentation{} on Spider}
To prepare the training data for NL2ERM models, we apply the \dataaugmentation{} algorithm on the training dataset of Spider \cite{yu-etal-2018-spider}, which is a widely used cross-domain benchmark for the text-to-SQL task. 
The \dataaugmentation{} algorithm generates 140 ER models and 7000 NL utterances with token-level annotations, and each ER model corresponds to 50 NL utterances on average.

To evaluate the generalization ability of NL2ERM models, we conduct experiments on different types of NLs.
The test dataset consists of three parts, each of which contains ER models and a different type of NL: the question, software requirement, and scenario description:
\begin{itemize}
\item \textbf{Question}. We generate the part of the question test dataset by applying the \dataaugmentation{} algorithm to the validation dataset of Spider. The NL utterances in Spider are usually the type of question.
This part contains 20 ER models with 1034 NL utterances.
\item \textbf{Software Requirement}. The part of the software requirement test dataset contains two requirements (18 NL utterances in total): the requirement for the Banking Transaction System (BTS) and the requirement for the Library Management System (LMS), both of which  are widely used in NL2ERM research \cite{Novel-Natural-Language, Generating-ER-Diagrams}.
\item \textbf{Scenario Description}. The part of the scenario description test dataset contains 6 scenario descriptions (37 NL utterances in total) and their corresponding ER models. We follow the work of \citet{Scenario-based} to select these scenario descriptions and ER models from two well-known database books \cite{book_fundamental,book_database}.
%\item The test dataset of user story xxxxx
\end{itemize}
The statistical results of our dataset are shown in Table \ref{tab:dataset}.

\subsection{Example Case}
\begin{figure}
\centering 
\subfigure[Some $<$NL utterance, SQL query$>$ pairs. The tokens with the label \textit{Table} or \textit{Column} are underlined in yellow and blue. The relationships are labeled as arrows.]{
\label{fig.case.1}
\includegraphics[width=0.45\textwidth]{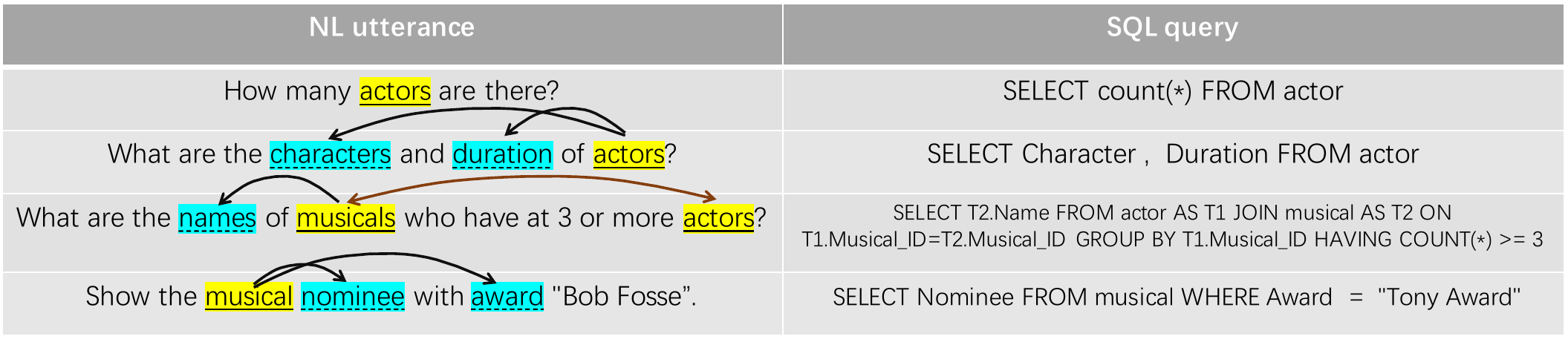}}
\subfigure[The database queried by the  SQLs in (a); our algorithm transform it into an ER model.]{
\label{fig.case.2}
\includegraphics[width=0.45\textwidth]{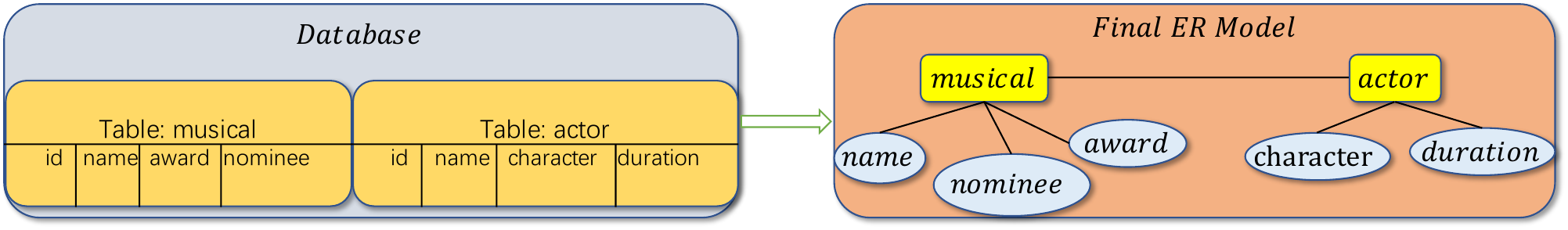}}
\caption{A database and the corresponding $<$NL utterance, SQL query$>$ pairs in Spider, where we simplify the database and show the final ER model.}
\label{fig:case}
\end{figure}
In this section, we discuss an example case to illustrate the result of our \dataaugmentation{} algorithm.

As shown in Figure \ref{fig.case.1}, the NL utterances are annotated by schema linking with their corresponding SQL queries.
The tables/columns that appear in SQL queries are matched with n-grams in utterances to get the labels for each token. 
If a column belongs to a table, or there is a table joining between two tables in SQL queries, we treat them as an entity-attribute relationship and an entity-entity relationship, respectively. 
Finally, we can get a token-level annotation that can be used to train deep-learning-based models.
In the third utterance, there is an attribute with two entities.
In the fourth utterance, there is no conjunction such as ``of'' between the attribute ``nominee'' and the entity ``musical''.
In such cases, it is difficult for rule-based approaches to correctly distinguish the entity-attribute relationships.

As shown in Figure \ref{fig.case.2}, some of the columns do not appear in any preceding utterances, so the final ER model prunes these columns (attributes).
\section{Experimental Setup}
In this section, we show the models that we use for NL2ERM, the evaluation metric, and the baselines. 
%\footnote{https://github.com/lzw-pku/NL2ERM-data-transformation} 
\subsection{Model Training}
To attain  NL2ERM models, we train two state-of-the-art IE models named UIE \cite{uie} and REBEL \cite{rebel} on our generated NL2ERM dataset.
We treat both entities and attributes in the NL2ERM dataset as entities with entity types \textit{Entity} and \textit{Attribute} in IE.
We also treat relationships in the NL2ERM dataset as relationships with two relation types ``entity-entity'' and ``entity-attribute''.
In this way, our NL2ERM dataset can be seen as a special IE dataset, and the IE models UIE and REBEL can be trained on it.
\subsection{Evaluation}
The metrics to evaluate the performance of NL2ERM  approaches are precision, recall, and $F_1$-score.
Considering that for each data entry, both the ground truth ER model and the inference ER model consist of three sets: the entity set, attribute set, and relationship set, we evaluate the three sets of the ER models separately.
Given the ground truth entity set $E_g$ and the inference entity set $E_i$, we compute the metrics as follows:
\begin{equation}
\begin{aligned}
&Precision = \frac{|E_g\cap E_i|}{|E_i|} \\
&Recall = \frac{|E_g\cap E_i|}{|E_g|} \\
&F_1 = \frac{2\times Precision\times Recall}{Precision + Recall}
\end{aligned}
\end{equation}
where $|\cdot|$ represents the size of a set.
The evaluation for the attribute set and relationship set is the same way.
\subsection{Implementation}
During \dataaugmentation{}, we use the spaCy package \cite{spacy} to preprocess NL utterances.

We train the UIE model on our training dataset according to the UIE authors' script.
We train it for 50 epochs, on a 3080 GPU for six hours.
We use the Adam optimizer \cite{adam} with a learning rate of 1e-4.
The training batch size is 16.

We also train the REBEL according to the REBEL authors' script.
We train it for 50 epochs too, on a 3080 GPU for five hours.
We also use the Adam optimizer \cite{adam} with a learning rate of 1e-4.
The training batch size is 16.

%We train the two modules of the deep learning model (the entity/attribute extractor and the relationship predictor) separately.
%We train the model in a 3080 GPU within one hour.
%Our model has a total parameter of 221M.
%Both modules use BERT \cite{devlin-etal-2019-bert} to encode the input sentences.
%For the entity/attribute extractor, the training batch size is 16, and the number of epochs is 4. 
%The extractor uses the AdamW optimizer \cite{adamw} with a learning rate of 1e-5.
%For the relationship predictor, the training batch size is 32, and the number of epochs is 4. 
%The predictor also uses the AdamW optimizer with the same learning rate.
\subsection{Baselines}
Because the authors of the previous NL2ERM approaches~\cite{Scenario-based, Novel-Natural-Language, Generating-ER-Diagrams} do not release their corresonding tools and do not respond to our requests for their tools, we reproduce/reimplement two very recent rule-based tools.

\textbf{ER-Gen}. \citet{Novel-Natural-Language} propose a rule-based tool that relies on the help of the OpenNLP API. We reproduce this tool because it is the latest work in this field, so it can represent  one of the state-of-the-art approaches  in this field.

\textbf{Scenario-ER}. \citet{Scenario-based} propose a rule-based tool that can extract an ER model from a given scenario. The tool uses six, four, and one rules for entity, attribute, and relationship extraction, respectively. We reproduce this tool  because the heuristic rules used in this tool are also widely used in previous work \cite{Generating-ER-Diagrams}.
\begin{table*}[!t]
  \centering
    \begin{tabular}{l|ccc|ccc|ccc|ccc}
        \hline
        \multirow{2}*{\diagbox{Dataset}{Model}}&\multicolumn{3}{c}{UIE}&\multicolumn{3}{c}{REBEL}&\multicolumn{3}{c}{ER-Gen}&\multicolumn{3}{c}{Scenario-ER}\\
        \cline{2-13}
         ~&E&A&R&E&A&R&E&A&R&E&A&R\\
         \hline
         Question&0.76&0.78&0.55&0.78&0.78&0.48&0.49&0.34&0.06&0.48&0.35&0.08 \\
         Requirement&0.87&0.86&0.69&0.77&0.82&0.60&0.75&0.78&0.14&0.73&0.81&0.21 \\
         Scenario&0.79&0.71&0.52&0.72&0.71&0.54&0.69&0.63&0.10&0.61&0.58&0.11 \\
        \hline
    \end{tabular}
    \caption{Main experimental results, where the scores in the table are $F_1$ scores, and E, A, and R represent entity, attribute, and relationship extraction, respectively.}
    \label{tab:main result}
\end{table*}
\section{Results and Discussion}
\subsection{Main Results}
As shown in Table \ref{tab:main result}, two deep-learning-based models achieve high $F_1$-scores and outperform all the baselines in all three parts of the test dataset.
%The experimental results show that 
% our deep learning model achieves high performance and outperforms the rule-based baseline for the NL2ERM task. 

Among the three parts of the test dataset, the question part is the most difficult for both the deep-learning-based models and the baselines.
A reason is that the formats of the questions in Spider vary substantially, while the formats of software requirements and scenario descriptions are relatively fixed. Despite the difficulty, our model still achieves high performance (around 80\% F-1 score on entity/attribute) whereas the baselines achieve low performance. 
%REBEL performs worse than our model because the data distribution is different between the IE and NL2ERM datasets (``part of'' and ``has part'' relationships are only a small fraction of all relationships).
%This result shows that IE models cannot be directly used for NL2ERM. 

Although an NL utterance in the training dataset is of the type of question, the deep-learning-based models achieve a high $F_1$ score in the parts of software requirements and scenario descriptions.
The success illustrates the effectiveness of our \dataaugmentation{} algorithm.
The main reason for the success is that both questions and other types of NL utterances contain similar and sufficient information about the ER model, and the consistency in information helps the model to overcome the gap between different types of NL utterances in the training and test datasets.

\subsection{Ablation on Generated Dataset}
\begin{table}[!t]
  \centering
    \begin{tabular}{l|ccc|ccc}
        \hline
        ~&\multicolumn{3}{c}{Requirement}&\multicolumn{3}{c}{Scenario}\\
        \cline{2-7}
        ~&E&A&R&E&A&R\\
        \hline
         UIE&0.87&0.86&0.69&0.79&0.71&0.52\\
        w/o &0.35&0.10&0.04&0.26&0.08&0.02\\
        \hline
        REBEL&0.77&0.82&0.60&0.72&0.71&0.54\\
         w/o &0.47&0.39&0.25&0.25&0.36&0.05\\
        \hline
    \end{tabular}
    \caption{The results of the ablation study, where  
    ``w/o'' denotes the model without training on our dataset, and the scores in the table are $F_1$ scores. For a fair comparison, the models are evaluated on only the requirement and scenario parts of the test dataset.}
    \label{tab: ablation study}
\end{table}
To investigate the effect of our generated training dataset, we also conduct an ablation study by evaluating the IE models without training on our dataset. 
We train UIE and REBEL on some other IE datasets, according to their authors'  scripts.
We train UIE on the NYT dataset \cite{nyt} and REBEL on the REBEL dataset \cite{rebel}.

When evaluating the IE models without training on our 
 NL2ERM dataset, there are two difficulties: (1) the outputs of IE models are $<entity_1, relation, entity_2>$, and we need to distinguish entities and attributes; (2) the set of relation types is different between the IE training datasets and our test dataset.

To overcome the difficulties, we choose some special relation types as entity-attribute relationships, based on the semantics of the relation types.
An entity-attribute relationship indicates that the two entities contained in the relationship can be seen as an entity and an attribute in an ER model.
According to the IE datasets (NYT and REBEL) and our observation on models outputs, we choose the relation types \textit{contains} as the entity-attribute relationship for UIE and \textit{part of}, \textit{has part}, \textit{subclass of
}, and \textit{facet of} as entity-attribute relationships for REBEL.
Given a model output, we can first distinguish entities and attributes by these entity-attribute relationships, and then we treat all remaining relationships as entity-entity relationships in the ER model.

As shown in Table \ref{tab: ablation study}, the IE models trained on our generated dataset perform better, and the results illustrate our dataset is essential.
\begin{table}[!t]
  \centering
    \begin{tabular}{l|ccc}
        \hline
        ~&P&R&$F_1$\\
         \hline
         Entity&0.90&0.94&0.92 \\
         Attribute&0.96&0.96&0.96 \\
         Relationship&0.92&0.91&0.92 \\
        \hline
    \end{tabular}
    \caption{Human evaluation result for the dataset, where P, R, and $F_1$ are precision, recall, and $F_1$ score,  respectively.}
    \label{tab:human_eval}
\end{table}
%\begin{figure}[!t]
% \centering  %图⽚全局居中
 %\subfigbottomskip=2pt %两⾏⼦图之间的⾏间距
 %\subfigcapskip=-5pt %设置⼦图与⼦标题之间的距离
% \subfigure[The software requirement for the LMS.]{
%  \includegraphics[width=1\linewidth]{EMNLP 2022/image/case_study_a.pdf}
%  \label{fig:case_study_a}}
%  \\
% \subfigure[A query and a scenario description.]{
%  \includegraphics[width=1\linewidth]{EMNLP 2022/image/case_study_b.pdf}
%  \label{fig:case_study_b}}
%   \\
% \subfigure[A sentence from a scenario description.]{
%  \includegraphics[width=1\linewidth]{EMNLP 2022/image/case_study_c.pdf}
%  \label{fig:case_study_c}}
%  \caption{Three example cases (all of which are from our test dataset and labeled with ground truth).  
%The ground truth entities are labeled yellow with a solid underline, and the ground truth attributes are labeled blue with a dashed underline.
%In Figure \ref{fig:case_study_c}, the relationships between the entities and attributes are labeled as arrows.}
%\end{figure}
%\subsection{Comparison with Information Extraction (IE) Model}
%For further comparison, we choose a information extraction (IE) model named REBEL \cite{rebel}, which is one of the state-of-the-art model in this field.
%IE 

\subsection{Dataset Evaluation}
To evaluate the annotation accuracy of our \dataaugmentation{} algorithm, we randomly select 50 NL utterances from Spider. 
We annotate the table names, column names, and relationships from the NL utterances.
We compare the manual annotation result with our automatic schema linking result, and Table \ref{tab:human_eval} shows the evaluation results.
The results show that our \dataaugmentation{} algorithm achieves a high $F_1$ score on the entity, attribute, and relationship extraction, and our generated dataset is of  sufficient quality to be used to train NL2ERM models.
%
%During dataset transformation, the schema linking stage links NL tokens with table/column names of 
%When transforming NL2SQL datasets into NL2ERM datasets, our \dataaugmentation{} algorithm 
%We also measure the coverage of our \dataaugmentation{} algorithm.
%
\begin{figure}[!t]
\centering 
\includegraphics[width=1\linewidth]{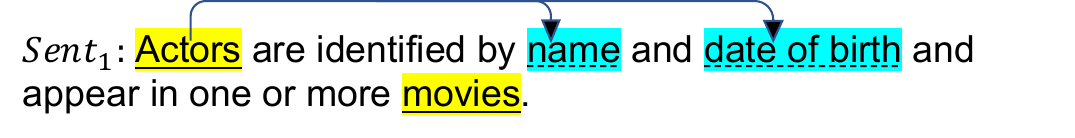}
\caption{A sentence from a scenario description, where the ground truth entities and attributes are underlined in yellow and blue, and the relationships are labeled as arrows.}
\label{fig:case_study_c}
\end{figure}
\subsection{Example Case}
We show an example case to illustrate the advantages of deep-learning-based models compared with existing rule-based approaches.
%
%For entity/attribute extraction, the rule-based approaches treat the nouns in the sentences as entities or attributes.
%One difficulty is how to distinguish entities and attributes.
%A widely used rule is that \textit{the consecutive nouns separated by commas or ``and'' are attributes.} \cite{Novel-Natural-Language, Scenario-based}.
%As shown in Figure \ref{fig:case_study_a}, $sent_1$ and  $sent_4$ can cause challenges for the rule-based baseline.
%$Sent_1$ is a sentence that describes the whole system, so the consecutive nouns separated by comma or ``and'' are entities, and these nouns are wrongly identified as attributes by the baseline.
%In $Sent_4$, the entity ``author'' has only one attribute ``author name'', and the attribute is not separated by commas or ``and'', so  $Sent_4$ also causes challenges for  the baseline.
%In contrast, our deep learning model successfully distinguishes the entities and attributes in this case. 
%
%Another disadvantage of the rule-based approaches is that all the common nouns are identified as entities or attributes \cite{heuristic-based, Generating-ER-Diagrams, Novel-Natural-Language}.
%As shown in Figure \ref{fig:case_study_b}, both words ``order'' in $sent_1$ and ``minutes'' in $sent_2$ are nouns and are wrongly identified as an entity by the rule-based baseline.
%In contrast, our deep learning model successfully excludes these two nouns because the model identifies entities/attributes based on the context and semantics, instead of rigid rules.
%
For relationship extraction, some rule-based approaches use a rule: \textit{the attributes must belong to the entity in the same sentence} \cite{Novel-Natural-Language, Scenario-based, heuristic-based}.
This rule works based on the  assumption that there is no sentence containing attributes or two or more entities, while in some compound sentences, the assumption does not hold.
As shown in Figure \ref{fig:case_study_c}, there are two entities ``actor'' and ``author'' in a sentence, with the attributes ``name'' and ``date of birth''. In this case, the UIE model identifies the entities/attributes and determines that both attributes belong to the entity ``actor'', while the baselines wrongly determine  that the attributes belong to both entities ``actor'' and ``movie'' based on the rule.
The failure of rule-based approaches shows that due to the various linguistic ways of describing the same requirement, the generalization ability of rule-based approaches is poor.
\section{Conclusion}
In this paper, we have proposed a \dataaugmentation{} algorithm that can automatically transform text-to-SQL datasets to NL2ERM datasets. 
We have applied  the algorithm on Spider, one of the most popular text-to-SQL datasets, to produce a large-scale NL2ERM dataset, which can be used to train deep-learning-based models for NL2ERM.
We have collected software requirements, scenarios descriptions, and their corresponding ER models to complete the test dataset.
To the best of our knowledge, we are the first to collect and release a large-scale dataset of NL2ERM.
We have also trained two state-of-the-art IE models on our training dataset and evaluate them.
The results show that our large-scale dataset is helpful for model training, and the two deep-learning-based models can achieve high performance on our test dataset.
\section{Limitation}
Our work has three major limitations.
First, the number of software requirements in the test dataset is small, because we can collect only two publicly released software requirements (BTS and LMS).
Second, the two deep-learning-based models are not directly compared with previous NL2ERM work's rule-based approaches. The authors of the previous rule-based approaches do not release their corresponding tools, and do not respond to our requests of their tools; thus, we reproduce/reimplement two very recent rule-based tools~\cite{Novel-Natural-Language,Scenario-based} as baselines compared in our experiments.
Third, the two deep-learning-based models cannot extract the cardinality of the relationships due to lacking  annotations on cardinality.
\bibliography{aaai24}

\end{document}